\documentclass[letterpaper, 10 pt, conference]{ieeeconf}  

\IEEEoverridecommandlockouts                              

\usepackage{times} 
\usepackage{amsmath} 
\usepackage{amssymb}  
\usepackage{url}
\usepackage{wrapfig}
\usepackage{booktabs}
\usepackage{makecell}
\usepackage{graphicx}
\usepackage{subcaption} 

\usepackage{xcolor} 

\newcommand{\revise}[1]{#1}

\title{\LARGE \bf
Compose by Focus: Scene Graph-based Atomic Skills \vspace{-2mm}}

\author{
Han Qi$^{1}$, 
Changhe Chen$^{2}$,
and Heng Yang$^{1}$ \\[2mm]
$^1$School of Engineering and Applied Sciences, Harvard University \\
$^2$Robotics Department, University of Michigan \\[1mm]
\url{https://computationalrobotics.seas.harvard.edu/SkillComposition/}
}

\begin{document}

\maketitle


\begin{abstract}
	

	\revise{A key requirement for generalist robots is compositional generalization—the ability to combine atomic skills to solve complex, long-horizon tasks. While prior work has primarily focused on synthesizing a planner that sequences pre-learned skills, robust execution of the individual skills themselves remains challenging, as visuomotor policies often fail under distribution shifts induced by scene composition. To address this, we introduce a \emph{scene graph-based} representation that \emph{focuses on} task-relevant objects and relations, thereby mitigating sensitivity to irrelevant variation. Building on this idea, we develop a scene-graph skill learning framework that integrates graph neural networks with diffusion-based imitation learning, and further combine ``focused'' scene-graph skills with a vision-language model (VLM) based task planner. Experiments in both simulation and real-world manipulation tasks demonstrate substantially higher success rates than state-of-the-art baselines, highlighting improved robustness and compositional generalization in long-horizon tasks.}

\end{abstract}

\section{Introduction}
\label{sec:introduction}
Solving long-horizon manipulation tasks requires breaking the whole task into multiple sub-tasks and executing individual skills (i.e., manipulation primitives). While much prior work has focused on how to compose \revise{already} learned skills (e.g., by learning a high-level planner) \cite{shah2021value, mao2024robomatrix, mei2024replanvlm}, in this paper we address an orthogonal problem: how should the individual skills themselves be constructed so that they can be effectively composed? 

\begin{figure}[h]
	\centering
	\includegraphics[width=0.95\linewidth]{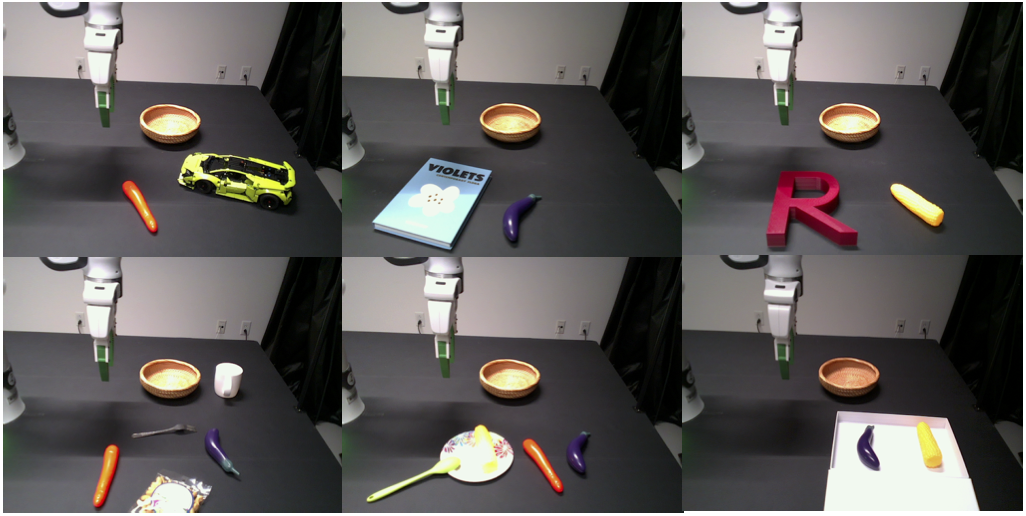}
	\caption{\textbf{Real world vegetable picking}. A policy trained to pick up a single vegetable on a clean table is evaluated by placing all vegetables into the basket in a cluttered scenario.
	\label{fig:realworld_veg}}
	\vspace{-10mm}
\end{figure}

\textbf{Motivating example.} Suppose the task is to pick up all the vegetables and put them into a basket. When presented with a cluttered scene containing many objects (including vegetables and distractors, as shown in Fig.~\ref{fig:realworld_veg}), a high-level planner—such as a Vision-Language Model (VLM) or Task and Motion Planner (TAMP) \cite{garrett21review-tamp}—can decompose the task into sub-goals: 1) pick up the carrot, 2) pick up the eggplant, 3) pick up the corn, etc. However, if each skill has only been trained in a plain environment (e.g., imitation learning from demonstrations with a single object on a clean table), this policy often fails in cluttered settings (see evaluation results in \S\ref{sec:simulation_exp}). The failure is not due to the planner but arises because the visuomotor policies lack robustness to distribution shifts. Prior studies \cite{haresh2024clevrskills} also find that state-of-the-art imitation learning methods struggle on compositional tasks, largely due to brittle visual processing. 

We argue that for skills to be composable, they must be \emph{focused}—attending only to scene elements relevant to the skill at hand while ignoring ``distractors''.

Towards this goal, we introduce a scene graph–based representation of visual data for efficient skill composition. Whereas most state-of-the-art approaches represent observations as raw RGB images \cite{chi2023diffusion, black2024pi0, kim2024openvla} or 3D point clouds \cite{ze20243d, ke20243d}, we instead transform visual input into dynamic semantic 3D scene graphs \cite{armeni20193d}. These graphs are \emph{relevant-object centric}—constructed to include only objects and relations pertinent to the current skill, thus filtering out irrelevant visual noise. Nodes encode 3D geometry and semantic features of objects, while edges dynamically capture inter-object relations inferred from multimodal cues. We build these graphs by leveraging visual foundation models (e.g., Grounded-SAM \cite{ren24arxiv-groundedsam}) for object segmentation and vision-language models (e.g., ChatGPT) for relation inference, yielding structured inputs that bypass raw image processing. We then apply Graph Neural Networks (GNNs) \cite{wu2020comprehensive} to extract graph features, which \revise{are then used to} condition diffusion-based visuomotor policies. This design substantially mitigates distribution shift in skill execution and enables robust composition of many atomic skills into long-horizon behaviors. In short, our method proposes \revise{using} \emph{scene graph} as an effective and interpretable way of data representation for \emph{skill composition}, improving the \emph{compositional generality} for solving long-horizon manipulation tasks.

\noindent \textbf{Contributions.} Our main contributions are:  

\begin{enumerate}
    \item We propose encoding structural scene graphs as general and interpretable inputs for vision-based policy learning using behavior cloning, where graphs capturing objects and relations are constructed with the help of VLMs and visual foundation models.  
    \item We integrate this representation with diffusion-based imitation learning and evaluate it on new skill composition manipulation benchmarks, showing substantial improvements over state-of-the-art baselines and strong robustness to visual perturbations in both simulation and real-world settings.  
\end{enumerate}

\section{Related Work}

\subsection{Compositional Generalization}

Skill composition is essential for generalization in robotics, as real-world tasks are long-horizon and multi-skilled. Decomposing tasks into sub-goals enables flexible recombination, and recent advances in language models \cite{du2023improving,subramaniam2025multiagent} have supported structured planning. Works combining LLMs with robotics \cite{mao2024robomatrix, mei2024replanvlm} focus on high-level planning via meta-skill libraries \cite{mao2023learning, zhu2022bottom}, but low-level execution remains underexplored. {Task and motion planning (TAMP)} \cite{garrett2021integrated, silver2021learning, curtis2022discovering} integrates discrete and continuous planning but depends on symbolic or geometric models, limiting generalization to complex or dynamic scenes. More recently, {object-centric methods} \cite{zhu2023learning, zhu2023viola} use VLMs for segmentation \cite{kirillov2023segment, cheng2022xmem} to build 3D features, improving over raw inputs but lacking explicit structure and broad generalization. In contrast, we propose scene graphs that encode objects and relations, enabling end-to-end skill composition and robust recomposition of atomic skills.

\subsection{Visual Imitation Learning}

Imitation learning is an efficient approach for training \revise{manipulation policies}, with visual imitation learning becoming \revise{increasingly popular} due to challenges in state estimation \revise{and reward engineering}. Most works \cite{chi2023diffusion, florence2022implicit, haldar2023teach, wang2023mimicplay, ha2023scaling, qi24icra-control} rely on \revise{2D images as policy input}, and robot foundation models similarly \revise{consume} 2D data \cite{black2024pi0, kim2024openvla}. While effective, 2D inputs lack spatial richness; 3D point-cloud policies \cite{ze20243d} and 4D temporal extensions \cite{niu2025pre} add structure but still \revise{lack explicit reasoning of objects and relations.} 

Beyond raw 2D/3D features, zero-shot generalization has been explored via disentangled latent spaces \cite{batra2025zero, hsu2024tripod}, but these often sacrifice interpretability. In contrast, we propose an explicit structural representation: scene graphs that encode objects and relations while embedding 3D information. This principled design supports robust skill composition by focusing on task-relevant context across varied environments.

\subsection{Scene Graphs in Robotics} 

Scene graphs have been widely explored as structured representations for perception and reasoning. Early work introduced 3D scene graphs as a way to unify semantics, geometry, and spatial relationships in visual scenes \cite{armeni20193d}. More recently, scene graphs have been used in robotics for high-level reasoning, such as task planning and navigation in cluttered or embodied environments \cite{singh2023scene, werby2024hierarchical}. However, these approaches typically employ scene graphs as auxiliary structures for planning or symbolic reasoning, rather than as direct inputs to low-level visuomotor policies. In contrast, our work explicitly encodes scene graphs into a behavior cloning framework, enabling robust skill composition under visual variation.

\section{Method}


\begin{figure*}[h]
	\centering
	\includegraphics[width=0.95\linewidth]{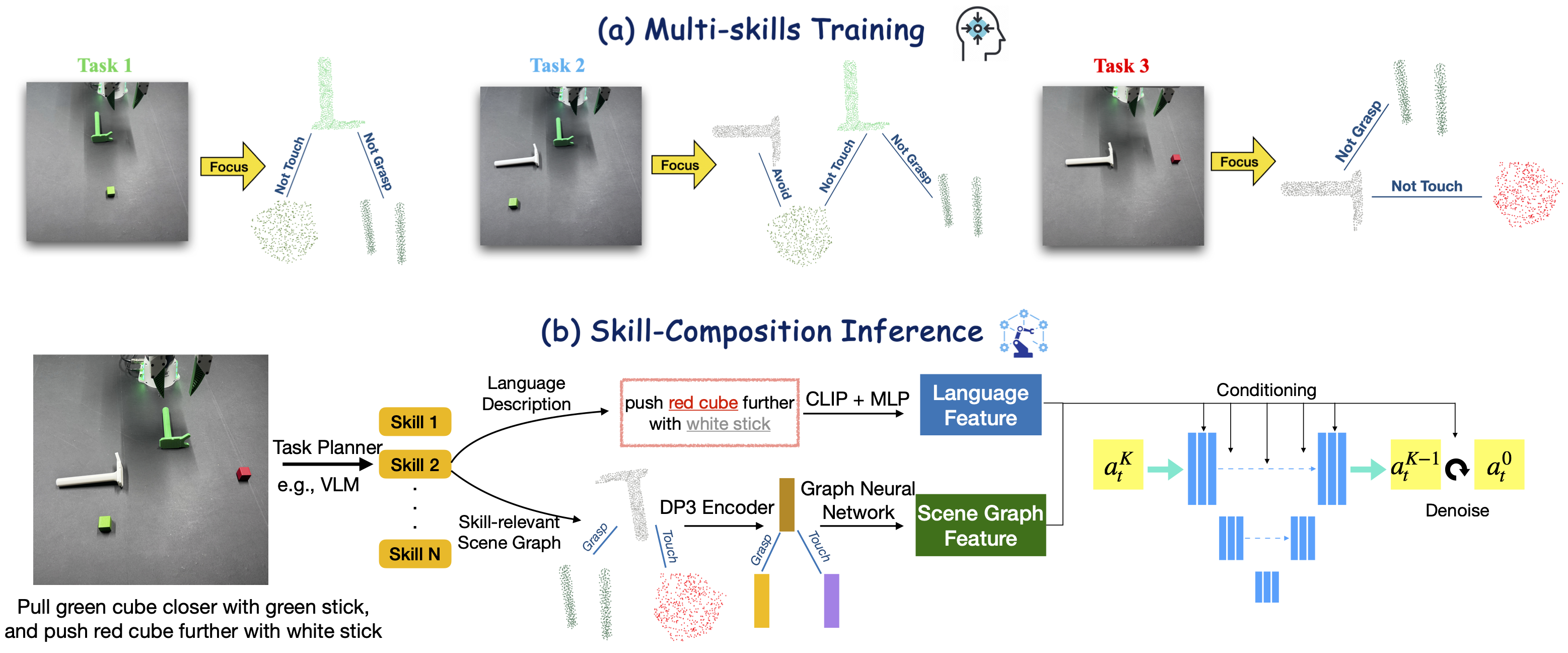}
	\caption{\textbf{Compose by focus}. (a) A single policy is trained on focused scene graphs across all sub-skills. (b) During inference, a task planner (e.g., VLM) decomposes a long-horizon task into $N$ sub-skills. For each sub-skill, CLIP encodes the description, Grounded SAM segments the relevant objects and extracts point clouds as graph nodes, and edges represent inter-object relations. DP3 Encoder \cite{ze20243d} embeds the nodes, a GNN encodes the scene graph, and a diffusion policy conditioned on both description and graph features iteratively denoises actions.
	\label{fig:main}}
	\vspace{-2mm}
\end{figure*}


\subsection{Problem Setup}

We address {long-horizon manipulation tasks} by composing atomic skills. Training data consist of expert demonstrations of individual skills, each manipulating objects toward a short-term goal (e.g., placing an apple in a bowl or pulling a cube with a tool). Unlike fixed primitives such as ``pick'' or ``place'' \cite{mishra2023generative, mishani2025mosaic}, our formulation defines atomic skills by task context. Importantly, demonstrations cover skills only in isolation, never their compositions in cluttered scenes.



Formally, we transform observations $O$ into scene graphs $G$ and have skill descriptions as $L$. A single visuomotor policy $\pi: (G, L) \mapsto A$ is trained to execute all atomic skills \revise{($A$ denotes the action space)}. The pipeline is illustrated in Fig.~\ref{fig:main}. We next describe (i) scene graph construction (\S\ref{method:scene_graph_representation}), (ii) policy learning on graph features (\S\ref{method:policy_learning}), and (iii) test-time skill composition (\S\ref{method:test_time_skill_composition}).

\subsection{Scene Graph Construction}
\label{method:scene_graph_representation}

\begin{figure}[h]
	\centering
	\includegraphics[width=0.6\linewidth]{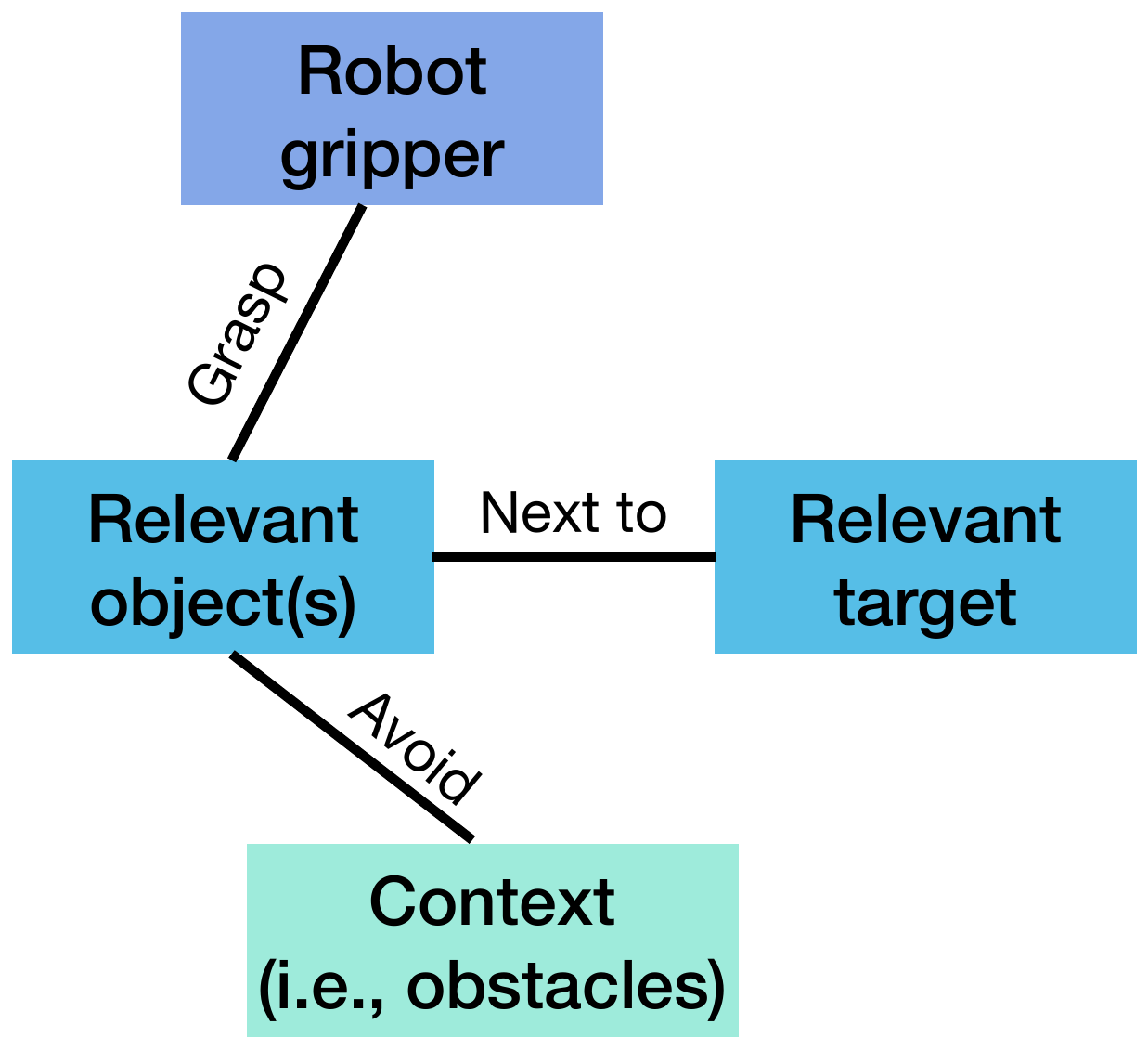}
	\caption{\textbf{Illustration of a scene graph}.
	\label{fig:scenegraph}}
	\vspace{-5mm}
\end{figure}

We obtain RGB and depth images from expert demonstrations in both simulation and real-world settings. Depth is converted to point clouds using camera intrinsics and extrinsics \cite{ze20243d}, and the robot gripper’s point cloud is assumed available. Each atomic skill is paired with a language description specifying the relevant objects $B$.  

To extract object-level information, we employ a vision foundation model (Grounded SAM \cite{ren24arxiv-groundedsam}) to segment masks of the \emph{task-relevant} objects from RGB images and obtain their corresponding point clouds. These point clouds are then downsampled using farthest point sampling \cite{moenning2003fast} and encoded into compact vector representations via a lightweight MLP network, DP3 Encoder \cite{ze20243d}. Each object's embeddings serve as the nodes of the scene graph.

Edges capture dynamic inter-object relations (e.g., `grasp', `next', `inside'), inferred from RGB with a VLM (e.g., ChatGPT). The resulting subgraph focuses only on task-relevant entities—robot gripper, objects, target, and optional obstacles (Fig.~\ref{fig:scenegraph}). Training on such concise subgraphs enables robust composition of learned skills in complex scenarios.

\subsection{Multi-skill Policy Training}
\label{method:policy_learning}

After transforming every frame of observation data into a task relevant sub-scene-graph, we implement a two-layer Graph Attention Network (GAT) \cite{velivckovic2017graph} to transform the scene graph into feature embedding. 

Each node $i \in \mathcal{V}$ is initialized with an input feature 
$\mathbf{h}_i^{(0)} \in \mathbb{R}^{d_{\text{in}}}$. 
At each GAT layer $\ell$, node features are updated as
\[
\mathbf{h}_i^{(\ell+1)} = \sigma_\ell \!\left( \big\Vert_{m=1}^{H_\ell}
\sum_{j \in \mathcal{N}(i)} \alpha_{ij}^{(\ell,m)} \,\mathbf{W}^{(\ell,m)} \mathbf{h}_j^{(\ell)} \right),
\]
where $\alpha_{ij}^{(\ell,m)}$ denotes the attention coefficient between node $i$ and $j$ under head $m$, 
and $\mathbf{W}^{(\ell,m)}$ is the learnable weight matrix of head $m$. 
Here, $\big\Vert_{m=1}^{H_\ell}$ denotes the concatenation of the outputs from all $H_\ell$ attention heads. 
Formally, if each head produces $\mathbf{z}_i^{(\ell,m)} \in \mathbb{R}^d$, then
\[
\big\Vert_{m=1}^{H_\ell} \mathbf{z}_i^{(\ell,m)} 
= \big[ \mathbf{z}_i^{(\ell,1)}; \mathbf{z}_i^{(\ell,2)}; \ldots; \mathbf{z}_i^{(\ell,H_\ell)} \big] 
\in \mathbb{R}^{H_\ell d}.
\]


We use $H_0 = 4$ heads with concatenation and $\sigma_0=\mathrm{ELU}$ (Exponential Linear Unit) in layer 1, $H_1=1$ head without concatenation and $\sigma_1=\mathrm{Id}$ in layer 2, yielding node embeddings $\{\mathbf{h}_i^{(2)}\}$.  
Graph representation (global mean pool) is then obtained as
\[
\mathbf{F} = \frac{1}{|\mathcal{V}|} \sum_{i \in \mathcal{V}} \mathbf{h}_i^{(2)}.
\]

With the embedded feature vector, this structured representation is able to be integrated into the state-of-the-art imitation learning framework, Diffusion Policy \cite{chi2023diffusion}.

For the language description of every atomic skill, we encode the text with CLIP encoder \cite{radford2021learning}, getting the skill description feature $P$.

We implement a visuomotor policy with conditional denoising diffusion model \cite{chi2023diffusion} to learn these atomic skills, conditioning on scene graph features $\mathbf{F}$, skill description features $P$ and robot poses $Q$ during $T_o$ observation horizons and then denoises random Gaussian noises into actions $A_t$, containing $T_p$ action steps starting from time $t$. Concretely, starting from the Gaussian noise $A_t^K$, the denoising network $\epsilon_\theta$ performs K iterations to denoise the random noise $A_t^K$ into the noise-free action $A_t^0$,
\begin{equation}
	A_t^{k-1} = \alpha_k \left( A_t^k - \gamma_k \, \epsilon_\theta \left(A_t^k, k, \mathbf{F}, P, Q\right) \right) + \sigma_k \mathcal{N}(0, \mathbf{I}),
\end{equation}
where $\mathcal{N}(0, \mathbf{I})$ is Gaussian noise, $\alpha_k$, $\gamma_k$, and $\sigma_k$ depend on the noise scheduler. 

To train the denoising network $\epsilon_\theta$, we add the noise $\epsilon^k$ at the $k$-th iteration to action $A_t^0$ \cite{chi2023diffusion}, and the training objective is to predict the noise added to the original action:
\begin{equation}
    \label{equation:training_objective}
    \mathcal{L} = \mathrm{MSE}\!\left( \epsilon^k, \, \epsilon_\theta \!\left( \bar{\alpha}_k A_t^0 + \bar{\beta}_k \epsilon^k, \, k, \mathbf{F}, P, Q \right) \right).
\end{equation}
where $\bar{\alpha}_k$ and $\bar{\beta}_k$ are the noise scheduler. In implementation, we end-to-end train the point cloud encoder, graph encoder and diffusion model.


\subsection{Testing Time Skill Composition}
\label{method:test_time_skill_composition}

To \revise{solve a long-horizon task}, we can leverage VLM (i.e., ChatGPT-4V) for high-level planning. For each sub-goal $S$, the VLM is used to parse the instruction and identify the relevant objects. Given each observation, Grounded SAM segments the point clouds of these objects in visually complex scenes, while VLM infers their semantic relationships. Based on this information, a dynamic sub-scene graph corresponding to the sub-goal is constructed, and its features are encoded using a GNN. As illustrated in Fig.~\ref{fig:main} (b), the trained policy then predicts actions conditioned on the sub-scene graph features and the sub-goal description. We assume that all atomic skills required at test time have been \revise{learned} during training.

\section{Simulation Experiment}
\label{sec:simulation_exp}

In this section, we evaluate our approach in simulated multi-skill environments. The primary goal of this work is to demonstrate the effectiveness of using scene graphs as inputs for behavior cloning, with a particular emphasis on skill composition. Our experiments are designed to highlight two main points. First, we show that scene graph–based representations enable \emph{zero-shot generalization} in \emph{compositional scenes}, whereas policies trained on raw 2D or 3D visual inputs suffer from severe distribution shifts. Second, we conduct an ablation study to assess the importance of the scene graph representation in our framework, demonstrating that every component plays a critical role in enabling robust skill composition. Together, these experiments validate the effectiveness and necessity of our design choices for solving long-horizon manipulation tasks.

\begin{figure*}[h]
	\centering
	\includegraphics[width=0.9\linewidth]{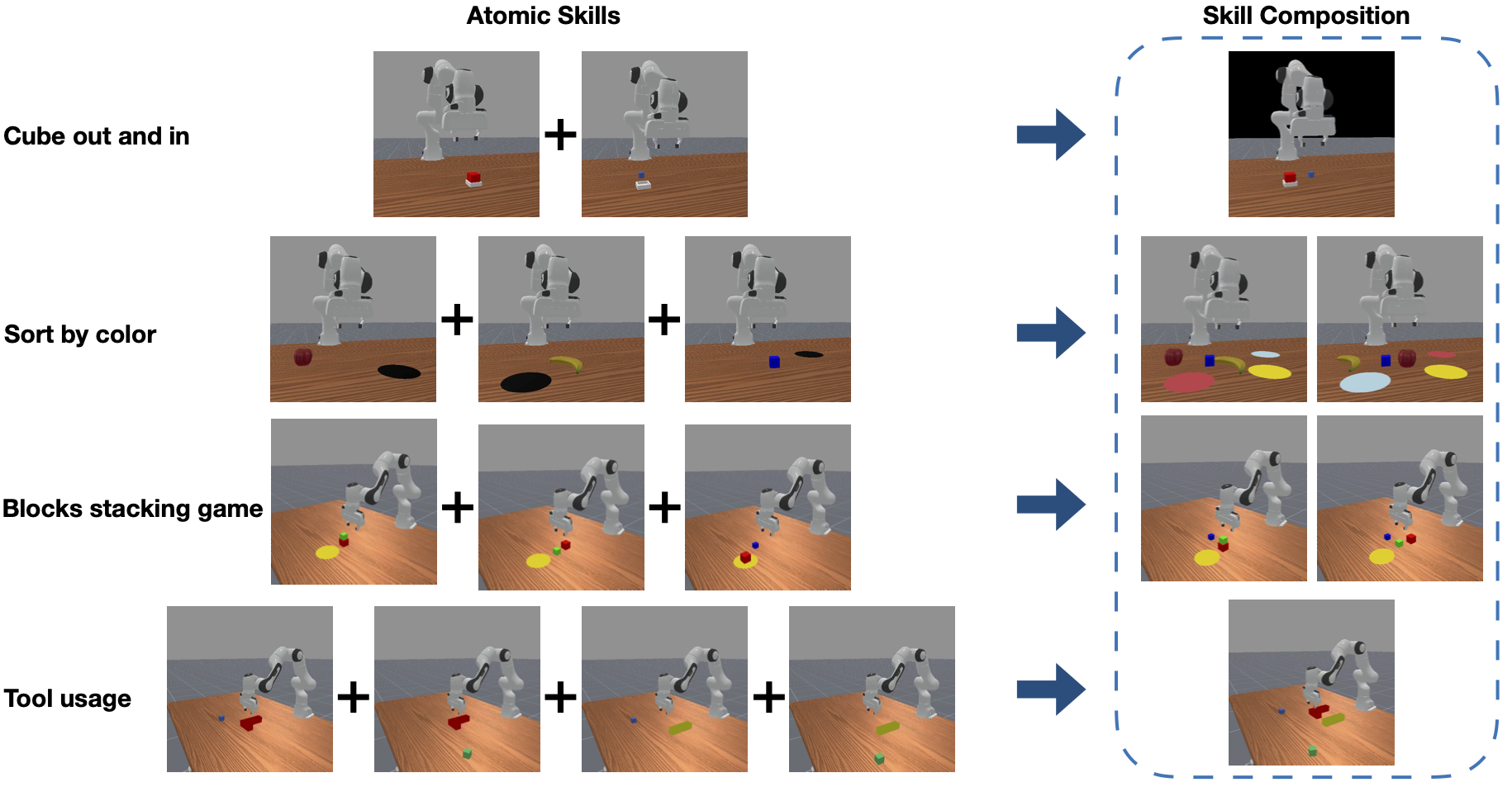}
	\caption{\textbf{Simulation tasks}. [Left] Visualization of atomic skills in each task, which is the training data. [Right] Evaluation scenarios, involving multiple objects to be operated with possibly changed background. 
	\label{fig:simulation-tasks}}
	\vspace{-2mm}
\end{figure*}

\begin{figure}[h]
	\centering
	\includegraphics[width=0.7\linewidth]{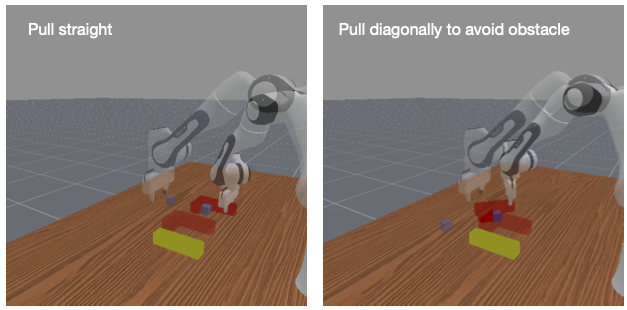}
	\caption{\textbf{Obstacle avoidance}. The sub-scene graph includes the robot gripper, relevant objects, and an \emph{obstacle node} (yellow stick), enabling the policy to learn trajectories that depend on the cube–obstacle relation.
		\label{fig:simulation-avoid}}
	\vspace{-4mm}
\end{figure}

\subsection{Task Description}

Since most existing simulation benchmarks (e.g., \cite{gu2023maniskill2, NVIDIA_Isaac_Sim}) focus primarily on single-skill tasks, we design five sets of multi-skill, long-horizon tasks built on ManiSkill2 \cite{gu2023maniskill2}, with illustrations provided in Fig.~\ref{fig:simulation-tasks}. These tasks cover 13 atomic skills spanning a range of manipulation primitives (e.g., pick, place, push, pull). Each multi-skill task is designed to require visual reasoning, generalization, and action understanding in order to compose atomic skills within visually complex scenes. To further increase difficulty, some decisions are conditioned on the scene—for example, whether the robot arm must avoid an obstacle when pulling a cube (see Fig.~\ref{fig:simulation-avoid}). We also provide motion-planning scripts for all atomic skills, which are used to generate expert demonstrations. For each atomic skill, we collect $100$ demonstrations. The tasks are described below:

\begin{itemize}
	\item \textbf{Cube out and in}: The atomic skills are \emph{removing the red cube from the bin} and \emph{putting the blue cube into the bin}. The evaluation scenario begins with a red cube inside the bin and a blue cube outside. The robot is asked to \emph{``put the blue cube into the bin so that only the blue cube remains inside.''}
	
	\item \textbf{Sort by color}: The atomic skills are to \emph{place an object (apple, banana, or cube) onto a target elliptical place}. In evaluation, three elliptical places with different colors are presented, and the robot is asked to \emph{``put the three objects onto the ellipses of same color.''}
	
	\item \textbf{Blocks stacking game}: The atomic skills involve logical operations on cubes, inspired by how humans naturally play with them: (a) if two cubes are stacked, push them to the center area together; (b) if the red cube is behind the green cube, lift the red cube before moving it to the center; (c) stack the purple cube on top of the red cube if it is empty on top. In evaluation, all three cubes appear together, and the robot is asked to \emph{``move the red cube to the center while avoiding the green cube, and then stack the purple cube on top of the red cube if there is no other cube on it''}. The policy must understand the logic underlying these atomic skills to successfully compose them.
	
	\item \textbf{Tools usage}: The atomic skills are \emph{using the L-shaped tool to pull a cube back} and \emph{using the stick to push a cube away}. In evaluation, both tools and two cubes are present, and the robot is asked to \emph{``pull back the blue cube with red tool and push away the green cube with yellow tool''} (or vice versa).
	
	\item \textbf{Obstacle avoidance}: This is a more challenging variant of tool usage. If pulling a cube straight back would result in a collision, the robot must instead pull it diagonally to avoid obstacles (see Fig.~\ref{fig:simulation-avoid}). The evaluation task is to \emph{``pull one cube back with red tool and push another cube away with yellow tool, while avoiding the obstacles.''}
\end{itemize}

\subsection{Baselines}
For baselines, we compare against diffusion policies with different visual representations, including the 2D Diffusion Policy \cite{chi2023diffusion} and the 3D Diffusion Policy \cite{ze20243d}. We also evaluate against $\pi0$ \cite{black2024pi0}, a foundation model trained on large-scale data, as a baseline for large-scale pretraining, and compare its performance with our scene graph–based approach on skill composition tasks. These comparisons motivate our hypothesis that a structured and focused representation of observations is essential for robust skill composition. It is important to note that our work emphasizes \emph{end-to-end} policy learning with scene graph representations. Although we leverage VLMs for high-level planning, they are only used to decompose abstract goals into concrete sub-steps. For a fair comparison, we also allow baselines to do the high-level planning at the beginning.

All methods are trained on the same expert demonstrations (RGB and depth images) and evaluated under the same maximum step limits. Each task is tested with $50$ randomized seeds (i.e., initial positions), and we report the average success rate across seeds. Each trial has a maximum success rate of $1.0$, defined as the proportion of completed sub-skills.

\subsection{Results}

\begin{table*}[h!]
\centering
\renewcommand{\arraystretch}{1.2}
\setlength{\tabcolsep}{6pt}

\begin{subtable}[t]{0.8\textwidth}
    \centering
    \begin{tabular}{lcccccc}
		\toprule
		Algorithm \textbackslash Task &
		Cube Out and In & Sort by Color & Blocks Stacking Game & Tools Usage & Obstacle Avoidance & \\
		\midrule
		
		Diffusion Policy & 
		$\mathbf{ 0.98 }$& 
		$\mathbf{ 1.0 }$& 
		$\mathbf{ 1.0 }$& 
		$\mathbf{ 1.0 }$& 
		$\mathbf{ 1.0 }$& \\
		
		DP3 & 
		$ 0.94 $& 
		$\mathbf{ 1.0 }$& 
		$ 0.96 $& 
		$ 0.92 $& 
		$\mathbf{ 1.0 }$& \\
		
		$\pi0$ & 
		$ 0.7 $& 
		$\mathbf{ 1.0 }$& 
		$ 0.94 $& 
		$ 0.26 $& 
		$\mathbf{ 1.0 }$& \\
		
		Scene Graph & 
		$\mathbf{ 0.98 }$& 
		$\mathbf{ 1.0 }$& 
		$ 0.94 $& 
		$\mathbf{ 1.0 }$& 
		$ 0.92 $& \\
		
		\bottomrule
	\end{tabular}
    \caption{Single Skill}
    \label{table:simulation_single_skill}
\end{subtable}
\hfill
\vskip 12pt
\begin{subtable}[t]{0.8\textwidth}
    \centering
    \begin{tabular}{lcccccc}
		\toprule
		Algorithm \textbackslash Task &
		Cube Out and In & Sort by Color & Blocks Stacking Game & Tools Usage & Obstacle Avoidance & \\
		\midrule
		
		Diffusion Policy & 
		$ 0.0 $& 
		$ 0.04 $& 
		$ 0.47 $& 
		$ 0.0 $& 
		$ 0.52 $& \\
		
		DP3 & 
		$ 0.27 $& 
		$ 0.07 $& 
		$ 0.48 $& 
		$ 0.13 $& 
		$ 0.44 $& \\
		
		$\pi0$ & 
		$ 0.15 $& 
		$ 0.02 $& 
		$ 0.77 $& 
		$ 0.07 $& 
		$ 0.49 $& \\
		
		Scene Graph & 
		$\mathbf{ 0.78 }$& 
		$\mathbf{ 0.79 }$& 
		$\mathbf{ 0.93 }$& 
		$\mathbf{ 0.88 }$& 
		$\mathbf{ 0.90 }$& \\
		
		\bottomrule
	\end{tabular}
    \caption{Skill Composition}
    \label{table:simulation_skill_composition}
\end{subtable}

\caption{\textbf{Evaluation results for all simulation tasks}. We evaluate in-domain performance on single-skill tasks (Table~(a)), where all methods achieve high scores. On composed tasks (Table~(b)), baseline performance drops by $50\%-70\%$ due to limited visual adaptation, while our method consistently outperforms them with higher and more stable success rates.}
\label{table:simulation_all}
\vspace{-4mm}
\end{table*}

\begin{figure*}[h]
	\centering
	\includegraphics[width=0.9\linewidth]{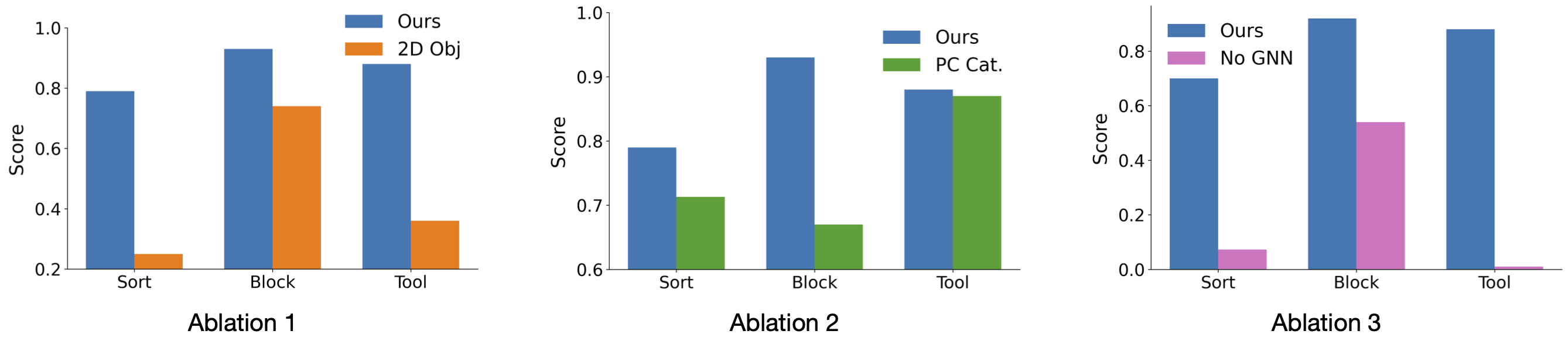}
	\vspace{-2mm}
	\caption{\textbf{Ablation studies highlighting the importance of 3D scene graph representations for skill composition}.
		\label{fig:ablation}}
	\vspace{-1mm}
\end{figure*}

\textbf{Evaluation on atomic skills:} Table~\ref{table:simulation_all}(a) shows that scene graph inputs enable near-perfect success ($\approx 1.0$) across all atomic skills, confirming their effectiveness for behavior cloning and the suitability of GNNs for encoding them. Additionally, the baselines also achieve good performance on single-skill tasks, confirming correct implementation and their ability to learn atomic skills. In contrast, as shown in the skill composition experiments of next part, the baselines fail to generalize when multiple skills must be composed.



\textbf{Evaluation on skill composition:} Next, we evaluate the policy on skill composition tasks. As shown in Table~\ref{table:simulation_all}(b), our composing-by-focus method yields only a small performance gap between atomic-skill tasks and compositional tasks. The policy consistently achieves high success rates across all compositional tasks. In contrast, the baseline methods experience substantial degradation: on average, their evaluation scores drop below $50\%$, and in some cases they fail completely. We want to emphasize a few key findings based on the results:
\begin{enumerate}
	\item \textbf{Sensitivity of behavior cloning to visual perturbations.} When additional objects are introduced into the scene or when the background changes, both 2D- and 3D-based policies often generate erroneous behaviors, failing to complete multi-step tasks.
	
	\item \textbf{Limitations of data scaling for skill composition.} The baseline $\pi0$, despite being pre-trained on large-scale robot manipulation datasets and further fine-tuned on our atomic-skill dataset, performs poorly on our skill composition experiments, indicating that it does not generalize well in these settings.


	\item \textbf{Challenges of domain adaptation.} For behavior cloning baselines, achieving skill composition would require demonstrations covering all permutations of atomic skills—an exponential growth relative to the number of skills, not mentioning the labor for collecting long-horizon demonstrations. In contrast, our approach enables composition directly from policies trained on atomic skills. The key insight is that VLMs/LLMs already demonstrate strong zero-shot reasoning, and scene graph representations provide a natural interface to integrate these capabilities with diffusion policies.  
	
	\item \textbf{Advantages of scene graph representations.} Scene graphs offer flexibility in the number of nodes and the information contained within them, while explicitly encoding inter-object relationships. This structure helps disambiguate subtle action differences (e.g., obstacle avoidance). 

\end{enumerate}

\subsection{Ablation Study}

We conduct three types of ablation studies to evaluate the importance of our 3D scene graph representation in training a policy with atomic skills for skill composition. While keeping high-level planning and object-centric representations fixed, we examine the contribution of specific design choices in our visual data processing pipeline. The ablations are performed on the ``Sort by Color'', ``Blocks Stacking Game'', and ``Tools Usage'' tasks.

\textbf{Importance of 3D representation}.
We ablate 3D object representations by cropping RGB images with Grounded SAM and extracting features using a pretrained DINOv2 encoder. Nodes are represented by these features while edges still encode inter-object relations. Trained with the same dataset and architecture, this 2D variant performs significantly worse, as shown in the left column of Fig.~\ref{fig:ablation}.


\textbf{Importance of graph representation}. A common alternative for 3D object-centric representations is to concatenate point clouds of related objects and encode them with a point cloud network (e.g., DP3 Encoder). While effective for simple single-skill tasks \cite{zhu2023learning}, this unstructured approach cannot handle varying object counts and weakens spatial relationships. Consequently, policies trained without scene graphs perform worse, as shown in Fig.~\ref{fig:ablation} (middle).

\textbf{Importance of GNN to process the graph}. Processing node features with a GNN captures inter-object relations, handles variable node counts, and is permutation-invariant. In contrast, concatenating node features discards edge information, fixes input size, and is sensitive to object order. An ablation with shuffled inputs (Fig.~\ref{fig:ablation}, right) shows our scene graph method remains robust, while the concatenation-based approach fails.

\section{Real World Experiment}

We evaluate the scene graph representation and skill composition in real-world across two multi-skill tasks, involving 6 atomic skills. We set up two RealSENSE L515 cameras to obtain real-world RGB images and depth images. Using intrinsics and extrinsics of the cameras, we are able to get the point clouds in world frame. We collect expert demonstrations by using a SpaceMouse, with $50$ demonstrations for each atomic skill.  For fair comparison, we give same number of maximum total steps for each evaluation among all the methods. We test $20$ trials for each task. Each trial has a maximum score of $1.0$, defined as the proportion of completed sub-skills (i.e., if there are three vegetables to pick and two are successfully picked, the score would be $0.67$).

\subsection{Real World Vegetable Picking}
\begin{table}[h]
	\centering
	\renewcommand{\arraystretch}{1.2}
	\resizebox{\linewidth}{!}{
	\begin{tabular}{lcccc}
		\toprule
		Task & Diffusion Policy & DP3 & $\pi0$ & Scene Graph \\
		\midrule
		Single Skill (w. distractors) & $ 0.6 $ & $ 0.7 $ & $ \mathbf{1.0} $ & $ \mathbf{1.0} $\\
		Skill Composition             & $ 0.0 $ & $ 0.2 $ & $ 0.05 $ & $ \mathbf{0.97} $\\
		\bottomrule
	\end{tabular}}
	\caption{\textbf{Real-world vegetable picking experiments}. Success rate is defined as the proportion of successfully picked vegetables. Compared with all baselines, our scene graph-based method is more robust to scenario variations and skill composition.}
	\label{table:realworld_vegetable}
	\vspace{-5mm}
\end{table}

		
		
		
		
		

\textbf{Single skill evaluation.}  
The atomic skill is to pick up a vegetable and place it in the basket. Training uses demonstrations of single vegetables (carrot/eggplant/corn) on a clean table, with scene graphs constructed via Grounded SAM \cite{ren24arxiv-groundedsam} containing three nodes: robot hand, vegetable, and basket. The diffusion policy is trained on these graph features and atomic skill descriptions.  

At evaluation, we test picking up a {single} vegetable in cluttered scenes (Fig.~\ref{fig:realworld_veg}, row 1). As shown in Table~\ref{table:realworld_vegetable}, our scene graph–based method achieves a perfect success rate ($1.0$), matching $\pi0$ but with far less pretraining. In contrast, 2D and 3D Diffusion Policies perform worse ($0.6$ and $0.7$), highlighting our method's robustness to distractors and data efficiency.



\textbf{Skill composition evaluation.}  
We then evaluate the policy's ability to compose multiple atomic skills. In this setting, several vegetables (with possible distractors) are placed on the table (see the second row of Fig.~\ref{fig:realworld_veg}), and the goal is described as \emph{``Pick up all vegetables and put them in the basket.''} To execute this goal, we use ChatGPT to generate substeps (e.g., \emph{``1. Pick up the corn and put it in the basket. 2. Pick up the carrot and put it in the basket.''}), and each sub-goal is passed sequentially to the policy. For each subtask, a sub-scene graph is constructed containing only the relevant objects, enabling the policy to focus on task-relevant context. An LLM is used to detect when a subtask finishes earlier than the maximum step limit, after which the next subtask begins.

As shown in Table~\ref{table:realworld_vegetable}, our method achieves a success rate of $0.97$ on skill composition, far outperforming the baselines: Diffusion Policy, DP3, and $\pi0$. While the baselines can learn atomic skills in isolation, they fail to generalize when multiple skills must be composed in cluttered scenes. These results demonstrate that scene graph–based representations are crucial for enabling robust multi-skill composition.

\subsection{Real World Tool Usage}

\begin{table}[h]
	\vspace{-2mm}
	\centering
	\renewcommand{\arraystretch}{1.2}
	\setlength{\tabcolsep}{8pt}
	\begin{tabular}{lcccc}
		\toprule
		Task & Diffusion Policy & DP3 & $\pi0$ & Scene Graph \\
		\midrule
		Tool Usage & $ 0.4 $ & $ 0.6 $ & $ 0.075 $ & $ \mathbf{0.9} $ \\
		\bottomrule
	\end{tabular}
	\caption{\textbf{Real-world tool usage experiments}. The score is the average score of $20$ trials. Each trial has a maximum score of $1.0$, defined as the proportion of completed sub-skills (i.e., pull cube or push cube).}
	\label{table:realworld_tool}
	\vspace{-3mm}
\end{table}

\begin{figure*}[h]
	\centering
	\includegraphics[width=0.85\linewidth]{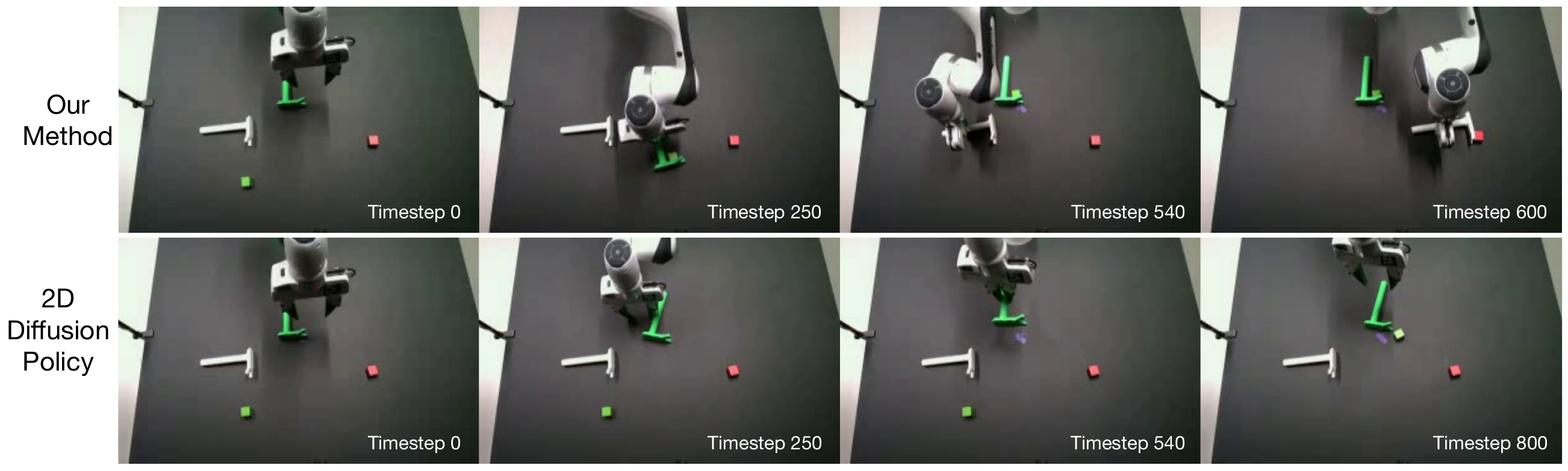}
	\caption{\textbf{Evaluation trajectories for a multi-skill task (pull then push with tools)}. The baseline (bottom) fails under visual complexity, while our method focuses on task-relevant objects via sub-scene-graphs and successfully composes the two skills. Additional trajectories are shown in the supplementary video.
	\label{fig:realworld_tool_traj}}
	\vspace{-2mm}
\end{figure*}

\begin{figure*}[h]
	\centering
	\includegraphics[width=0.85\linewidth]{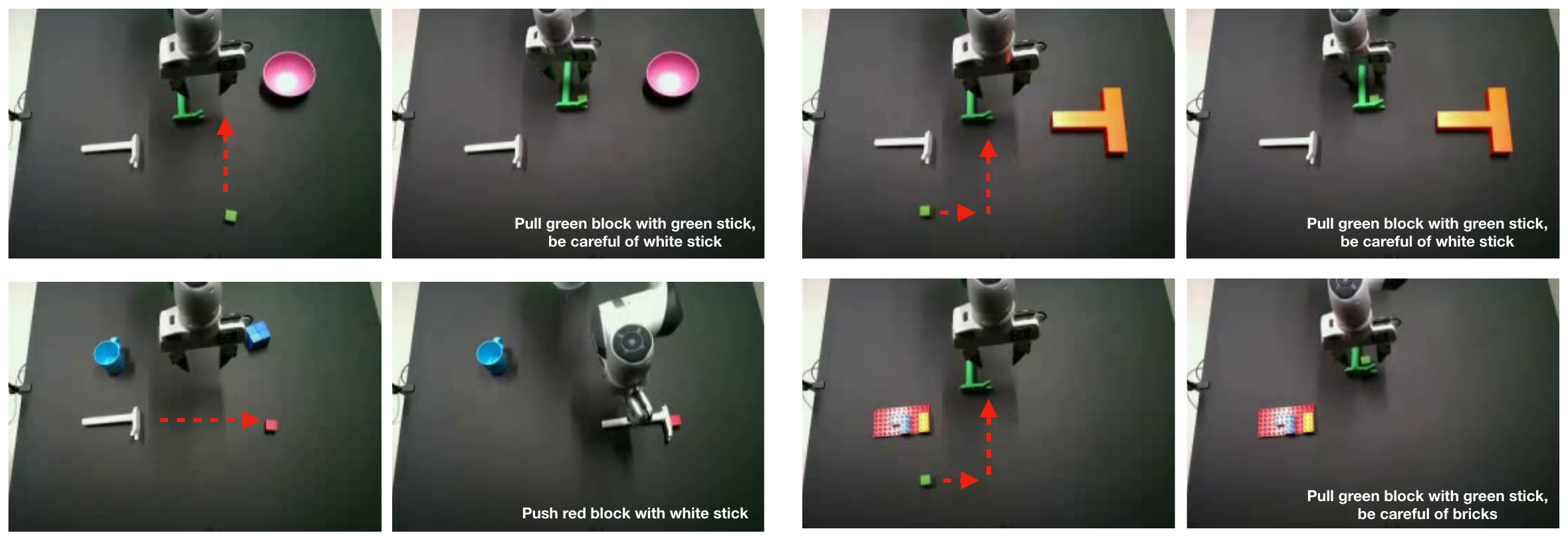}
	\caption{\textbf{Various evaluation settings for tool usage}. We test the generality for our scene graph-based methods by putting different distractors and obstacles on the table. Our method can give correct trajectories (like how to avoid the different obstacles) without being distracted by unrelated objects on the table. 
		\label{fig:realworld_tool_traj2}}
	\vspace{-4mm}
\end{figure*}

This experiment is the real-world counterpart of the tool-usage task, where an L-shaped stick is used to pull a cube closer or push it away, sometimes requiring obstacle avoidance (Fig.~\ref{fig:realworld_tool_traj2}). Atomic skills are defined as pulling or pushing a cube, with scene graphs containing nodes for `robot hand', `stick', `cube', and optional `obstacle'.  

For skill composition, two sticks and two cubes are placed on the table with the instruction: ``Use the green stick to pull the green cube, be careful of the white stick; then use the white stick to push the red cube.'' As in Table~\ref{table:realworld_tool}, our method, which focuses on task-relevant objects, achieves a $0.9$ success rate, far outperforming baselines. Example trajectories (Fig.~\ref{fig:realworld_tool_traj}) show baselines fail under visual variation and composition, while our approach composes skills reliably.  

Beyond skill composition, our approach also improves robustness. As shown in Fig.~\ref{fig:realworld_tool_traj2}, the policy (1) learns appropriate trajectories under different conditions (e.g., pulling straight versus pulling while avoiding an obstacle), and (2) remains resilient to scene variations such as background distractors or unseen obstacles. For instance, when the obstacle is changed from a stick (seen during training) to bricks (unseen), our method still achieves successful execution.

\section{Conclusions}
We introduced a scene graph–based representation for robotic skill composition, where nodes capture 3D information of task-relevant objects and edges encode their spatial relationships. This structured input enables visuomotor policies to focus on essential context, improving robustness to visual variation. By supporting flexible composition of atomic skills, our method reduces the exponential data demands of long-horizon demonstrations and improves data efficiency. Moreover, scene graphs provide a natural interface between high-level planning and low-level execution, offering a unified framework for skill composition by focus.


\section{Limitations}

Our method has two main limitations: (1) it relies on VLMs to dynamically construct scene graphs, which introduces additional computation overhead, though this remains manageable since we focus on a \emph{small} sub-scene graph; (2) it depends on the visual foundation model Grounded-SAM~\cite{ren24arxiv-groundedsam}, whose segmentation masks can occasionally be inaccurate, leading to errors. These issues are partly mitigated by providing precise descriptions, and we expect future improvements in VLMs and visual foundation models to further enhance both speed and accuracy of our pipeline.

\section*{ACKNOWLEDGMENT}
This work was partially funded by Office of Naval Research grant N00014-25-1-2322.


\bibliographystyle{IEEEtran}
\bibliography{refs.bib}

\begin{thebibliography}{10}
\providecommand{\url}[1]{#1}
\csname url@rmstyle\endcsname
\providecommand{\newblock}{\relax}
\providecommand{\bibinfo}[2]{#2}
\providecommand\BIBentrySTDinterwordspacing{\spaceskip=0pt\relax}
\providecommand\BIBentryALTinterwordstretchfactor{4}
\providecommand\BIBentryALTinterwordspacing{\spaceskip=\fontdimen2\font plus
\BIBentryALTinterwordstretchfactor\fontdimen3\font minus \fontdimen4\font\relax}
\providecommand\BIBforeignlanguage[2]{{%
\expandafter\ifx\csname l@#1\endcsname\relax
\typeout{** WARNING: IEEEtran.bst: No hyphenation pattern has been}%
\typeout{** loaded for the language `#1'. Using the pattern for}%
\typeout{** the default language instead.}%
\else
\language=\csname l@#1\endcsname
\fi
#2}}

\bibitem{shah2021value}
D.~Shah, P.~Xu, Y.~Lu, T.~Xiao, A.~Toshev, S.~Levine, and B.~Ichter, ``Value function spaces: Skill-centric state abstractions for long-horizon reasoning,'' \emph{arXiv preprint arXiv:2111.03189}, 2021.

\bibitem{mao2024robomatrix}
W.~Mao, W.~Zhong, Z.~Jiang, D.~Fang, Z.~Zhang, Z.~Lan, H.~Li, F.~Jia, T.~Wang, H.~Fan, \emph{et~al.}, ``Robomatrix: A skill-centric hierarchical framework for scalable robot task planning and execution in open-world,'' \emph{arXiv preprint arXiv:2412.00171}, 2024.

\bibitem{mei2024replanvlm}
A.~Mei, G.-N. Zhu, H.~Zhang, and Z.~Gan, ``Replanvlm: Replanning robotic tasks with visual language models,'' \emph{IEEE Robotics and Automation Letters}, 2024.

\bibitem{garrett21review-tamp}
C.~R. Garrett, R.~Chitnis, R.~Holladay, B.~Kim, T.~Silver, L.~P. Kaelbling, and T.~Lozano-P{\'e}rez, ``Integrated task and motion planning,'' \emph{Annual review of control, robotics, and autonomous systems}, vol.~4, no.~1, pp. 265--293, 2021.

\bibitem{haresh2024clevrskills}
S.~Haresh, D.~Dijkman, A.~Bhattacharyya, and R.~Memisevic, ``Clevrskills: Compositional language and visual reasoning in robotics,'' \emph{Advances in Neural Information Processing Systems}, vol.~37, pp. 38\,235--38\,266, 2024.

\bibitem{chi2023diffusion}
C.~Chi, Z.~Xu, S.~Feng, E.~Cousineau, Y.~Du, B.~Burchfiel, R.~Tedrake, and S.~Song, ``Diffusion policy: Visuomotor policy learning via action diffusion,'' \emph{The International Journal of Robotics Research}, p. 02783649241273668, 2023.

\bibitem{black2024pi0}
K.~Black, N.~Brown, D.~Driess, A.~Esmail, M.~Equi, C.~Finn, N.~Fusai, L.~Groom, K.~Hausman, B.~Ichter, \emph{et~al.}, ``$\pi_{0}$: A vision-language-action flow model for general robot control,'' \emph{arXiv preprint arXiv:2410.24164}, 2024.

\bibitem{kim2024openvla}
M.~J. Kim, K.~Pertsch, S.~Karamcheti, T.~Xiao, A.~Balakrishna, S.~Nair, R.~Rafailov, E.~Foster, G.~Lam, P.~Sanketi, \emph{et~al.}, ``Openvla: An open-source vision-language-action model,'' \emph{arXiv preprint arXiv:2406.09246}, 2024.

\bibitem{ze20243d}
Y.~Ze, G.~Zhang, K.~Zhang, C.~Hu, M.~Wang, and H.~Xu, ``3d diffusion policy: Generalizable visuomotor policy learning via simple 3d representations,'' \emph{arXiv preprint arXiv:2403.03954}, 2024.

\bibitem{ke20243d}
T.-W. Ke, N.~Gkanatsios, and K.~Fragkiadaki, ``3d diffuser actor: Policy diffusion with 3d scene representations,'' \emph{arXiv preprint arXiv:2402.10885}, 2024.

\bibitem{armeni20193d}
I.~Armeni, Z.-Y. He, J.~Gwak, A.~R. Zamir, M.~Fischer, J.~Malik, and S.~Savarese, ``3d scene graph: A structure for unified semantics, 3d space, and camera,'' in \emph{Proceedings of the IEEE/CVF international conference on computer vision}, 2019, pp. 5664--5673.

\bibitem{ren24arxiv-groundedsam}
T.~Ren, S.~Liu, A.~Zeng, J.~Lin, K.~Li, H.~Cao, J.~Chen, X.~Huang, Y.~Chen, F.~Yan, \emph{et~al.}, ``Grounded sam: Assembling open-world models for diverse visual tasks,'' \emph{arXiv preprint arXiv:2401.14159}, 2024.

\bibitem{wu2020comprehensive}
Z.~Wu, S.~Pan, F.~Chen, G.~Long, C.~Zhang, and P.~S. Yu, ``A comprehensive survey on graph neural networks,'' \emph{IEEE transactions on neural networks and learning systems}, vol.~32, no.~1, pp. 4--24, 2020.

\bibitem{du2023improving}
Y.~Du, S.~Li, A.~Torralba, J.~B. Tenenbaum, and I.~Mordatch, ``Improving factuality and reasoning in language models through multiagent debate,'' in \emph{Forty-first International Conference on Machine Learning}, 2023.

\bibitem{subramaniam2025multiagent}
V.~Subramaniam, Y.~Du, J.~B. Tenenbaum, A.~Torralba, S.~Li, and I.~Mordatch, ``Multiagent finetuning: Self improvement with diverse reasoning chains,'' \emph{arXiv preprint arXiv:2501.05707}, 2025.

\bibitem{mao2023learning}
J.~Mao, T.~Lozano-P{\'e}rez, J.~B. Tenenbaum, and L.~P. Kaelbling, ``Learning reusable manipulation strategies,'' in \emph{Conference on Robot Learning}.\hskip 1em plus 0.5em minus 0.4em\relax PMLR, 2023, pp. 1467--1483.

\bibitem{zhu2022bottom}
Y.~Zhu, P.~Stone, and Y.~Zhu, ``Bottom-up skill discovery from unsegmented demonstrations for long-horizon robot manipulation,'' \emph{IEEE Robotics and Automation Letters}, vol.~7, no.~2, pp. 4126--4133, 2022.

\bibitem{garrett2021integrated}
C.~R. Garrett, R.~Chitnis, R.~Holladay, B.~Kim, T.~Silver, L.~P. Kaelbling, and T.~Lozano-P{\'e}rez, ``Integrated task and motion planning,'' \emph{Annual review of control, robotics, and autonomous systems}, vol.~4, no.~1, pp. 265--293, 2021.

\bibitem{silver2021learning}
T.~Silver, R.~Chitnis, J.~Tenenbaum, L.~P. Kaelbling, and T.~Lozano-P{\'e}rez, ``Learning symbolic operators for task and motion planning,'' in \emph{2021 IEEE/RSJ International Conference on Intelligent Robots and Systems (IROS)}.\hskip 1em plus 0.5em minus 0.4em\relax IEEE, 2021, pp. 3182--3189.

\bibitem{curtis2022discovering}
A.~Curtis, T.~Silver, J.~B. Tenenbaum, T.~Lozano-P{\'e}rez, and L.~Kaelbling, ``Discovering state and action abstractions for generalized task and motion planning,'' in \emph{Proceedings of the AAAI conference on artificial intelligence}, vol.~36, no.~5, 2022, pp. 5377--5384.

\bibitem{zhu2023learning}
Y.~Zhu, Z.~Jiang, P.~Stone, and Y.~Zhu, ``Learning generalizable manipulation policies with object-centric 3d representations,'' \emph{arXiv preprint arXiv:2310.14386}, 2023.

\bibitem{zhu2023viola}
Y.~Zhu, A.~Joshi, P.~Stone, and Y.~Zhu, ``Viola: Imitation learning for vision-based manipulation with object proposal priors,'' in \emph{Conference on Robot Learning}.\hskip 1em plus 0.5em minus 0.4em\relax PMLR, 2023, pp. 1199--1210.

\bibitem{kirillov2023segment}
A.~Kirillov, E.~Mintun, N.~Ravi, H.~Mao, C.~Rolland, L.~Gustafson, T.~Xiao, S.~Whitehead, A.~C. Berg, W.-Y. Lo, \emph{et~al.}, ``Segment anything,'' in \emph{Proceedings of the IEEE/CVF international conference on computer vision}, 2023, pp. 4015--4026.

\bibitem{cheng2022xmem}
H.~K. Cheng and A.~G. Schwing, ``Xmem: Long-term video object segmentation with an atkinson-shiffrin memory model,'' in \emph{European conference on computer vision}.\hskip 1em plus 0.5em minus 0.4em\relax Springer, 2022, pp. 640--658.

\bibitem{florence2022implicit}
P.~Florence, C.~Lynch, A.~Zeng, O.~A. Ramirez, A.~Wahid, L.~Downs, A.~Wong, J.~Lee, I.~Mordatch, and J.~Tompson, ``Implicit behavioral cloning,'' in \emph{Conference on robot learning}.\hskip 1em plus 0.5em minus 0.4em\relax PMLR, 2022, pp. 158--168.

\bibitem{haldar2023teach}
S.~Haldar, J.~Pari, A.~Rai, and L.~Pinto, ``Teach a robot to fish: Versatile imitation from one minute of demonstrations,'' \emph{arXiv preprint arXiv:2303.01497}, 2023.

\bibitem{wang2023mimicplay}
C.~Wang, L.~Fan, J.~Sun, R.~Zhang, L.~Fei-Fei, D.~Xu, Y.~Zhu, and A.~Anandkumar, ``Mimicplay: Long-horizon imitation learning by watching human play,'' \emph{arXiv preprint arXiv:2302.12422}, 2023.

\bibitem{ha2023scaling}
H.~Ha, P.~Florence, and S.~Song, ``Scaling up and distilling down: Language-guided robot skill acquisition,'' in \emph{Conference on Robot Learning}.\hskip 1em plus 0.5em minus 0.4em\relax PMLR, 2023, pp. 3766--3777.

\bibitem{qi24icra-control}
H.~Qi, H.~Yin, and H.~Yang, ``Control-oriented clustering of visual latent representation,'' in \emph{International Conference on Learning Representations (ICLR)}, 2025.

\bibitem{niu2025pre}
D.~Niu, Y.~Sharma, H.~Xue, G.~Biamby, J.~Zhang, Z.~Ji, T.~Darrell, and R.~Herzig, ``Pre-training auto-regressive robotic models with 4d representations,'' \emph{arXiv preprint arXiv:2502.13142}, 2025.

\bibitem{batra2025zero}
S.~Batra and G.~Sukhatme, ``Zero-shot visual generalization in robot manipulation,'' \emph{arXiv preprint arXiv:2505.11719}, 2025.

\bibitem{hsu2024tripod}
K.~Hsu, J.~I. Hamid, K.~Burns, C.~Finn, and J.~Wu, ``Tripod: Three complementary inductive biases for disentangled representation learning,'' \emph{arXiv preprint arXiv:2404.10282}, 2024.

\bibitem{singh2023scene}
K.~P. Singh, J.~Salvador, L.~Weihs, and A.~Kembhavi, ``Scene graph contrastive learning for embodied navigation,'' in \emph{Proceedings of the IEEE/CVF International Conference on Computer Vision}, 2023, pp. 10\,884--10\,894.

\bibitem{werby2024hierarchical}
A.~Werby, C.~Huang, M.~B{\"u}chner, A.~Valada, and W.~Burgard, ``Hierarchical open-vocabulary 3d scene graphs for language-grounded robot navigation,'' in \emph{First Workshop on Vision-Language Models for Navigation and Manipulation at ICRA 2024}, 2024.

\bibitem{mishra2023generative}
U.~A. Mishra, S.~Xue, Y.~Chen, and D.~Xu, ``Generative skill chaining: Long-horizon skill planning with diffusion models,'' in \emph{Conference on Robot Learning}.\hskip 1em plus 0.5em minus 0.4em\relax PMLR, 2023, pp. 2905--2925.

\bibitem{mishani2025mosaic}
I.~Mishani, Y.~Shaoul, and M.~Likhachev, ``Mosaic: A skill-centric algorithmic framework for long-horizon manipulation planning,'' \emph{arXiv preprint arXiv:2504.16738}, 2025.

\bibitem{moenning2003fast}
C.~Moenning and N.~A. Dodgson, ``Fast marching farthest point sampling,'' University of Cambridge, Computer Laboratory, Tech. Rep., 2003.

\bibitem{velivckovic2017graph}
P.~Veli{\v{c}}kovi{\'c}, G.~Cucurull, A.~Casanova, A.~Romero, P.~Lio, and Y.~Bengio, ``Graph attention networks,'' \emph{arXiv preprint arXiv:1710.10903}, 2017.

\bibitem{radford2021learning}
A.~Radford, J.~W. Kim, C.~Hallacy, A.~Ramesh, G.~Goh, S.~Agarwal, G.~Sastry, A.~Askell, P.~Mishkin, J.~Clark, \emph{et~al.}, ``Learning transferable visual models from natural language supervision,'' in \emph{International conference on machine learning}.\hskip 1em plus 0.5em minus 0.4em\relax PmLR, 2021, pp. 8748--8763.

\bibitem{gu2023maniskill2}
J.~Gu, F.~Xiang, X.~Li, Z.~Ling, X.~Liu, T.~Mu, Y.~Tang, S.~Tao, X.~Wei, Y.~Yao, \emph{et~al.}, ``Maniskill2: A unified benchmark for generalizable manipulation skills,'' \emph{arXiv preprint arXiv:2302.04659}, 2023.

\bibitem{NVIDIA_Isaac_Sim}
\BIBentryALTinterwordspacing
{NVIDIA}, ``{Isaac Sim}.'' [Online]. Available: \url{https://github.com/isaac-sim/IsaacSim}
\BIBentrySTDinterwordspacing

\end{thebibliography}

\end{document}